\documentclass{article}

\usepackage[preprint]{corl_2026} 
\usepackage{amsmath}
\usepackage{amssymb}
\usepackage{graphicx}
\usepackage{subcaption}

\title{Self-Refining VLA Sampling}

\author{
  Jeff Duan \\
  Caltech \\
  \texttt{jxduan@caltech.edu} \\
  \And
  Arjun Sharma \\
  Caltech \\
  \texttt{arjun.sharma@caltech.edu} \\
  \AND
  Jennifer Cruden \\
  Caltech \\
  \texttt{jcruden@caltech.edu} \\
  \And
  Vansh Tibrewal \\
  Caltech \\
  \texttt{vansh@caltech.edu} \\
}

\begin{document}
\maketitle


\begin{abstract}
Vision-Language-Action (VLA) models typically use flow-based generative modeling to learn a rich, often multimodal, distribution of robot actions. Generating samples from flow-based methods involves iteratively updating random noise, and distinct noise samples from the same state can result in distinct  trajectories and task outcomes. Existing test-time scaling methods improve action chunk generation via selection, aggregation, or correction, but commonly require the generation of large ensembles of candidates or the training of an external corrector. We adapt Predict-and-Perturb (P\&P), a self-refinement procedure for flow-based models originally used for mode-seeking in video generation, to the robotics domain. We find that applying P\&P to VLA models requires modifications to the refinement process used in video generation due to differences in the underlying structure of the data. Across 400 episodes of LIBERO evaluated using SmolVLA, P\&P improves upon the stock baseline by $5.25\%$. Further, episode-level P\&P uncertainty correlates strongly with task failure (AUC $0.851$). 
\end{abstract}



\begin{figure}[htbp]
    \centering
    \begin{subfigure}[t]{\linewidth}
        \centering
        \includegraphics[width=\linewidth]{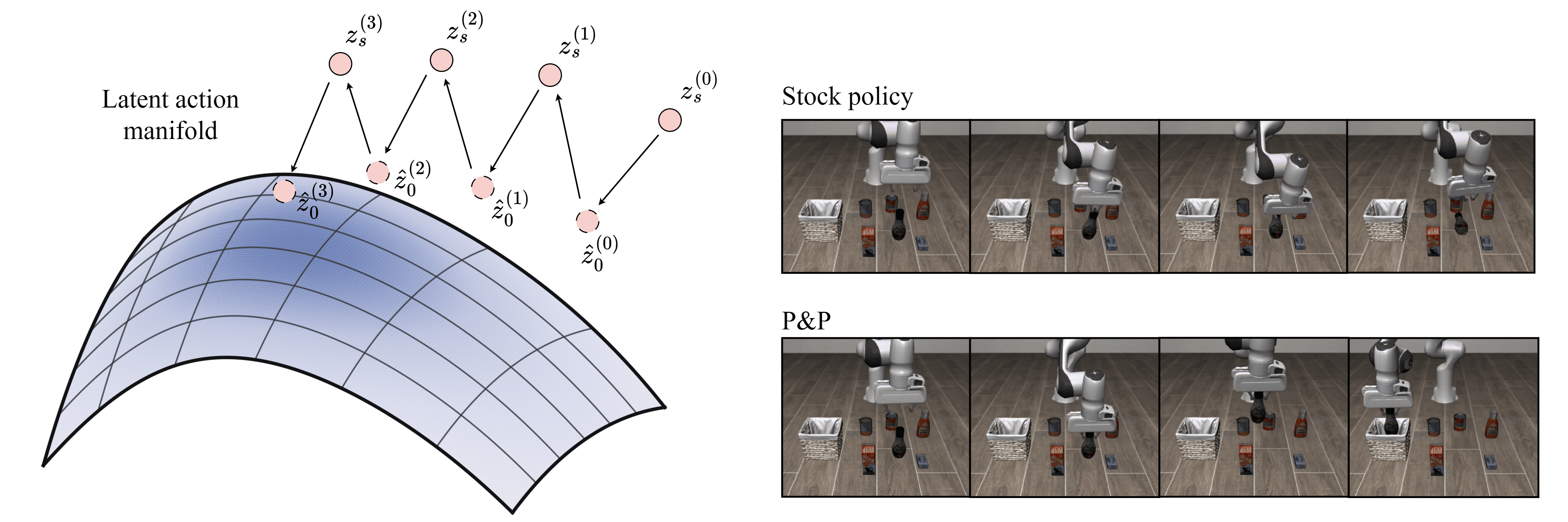}
        \label{fig:pnp-viz}
    \end{subfigure}\hfill
    \begin{subfigure}[t]{0.75\linewidth}
        \centering
        \includegraphics[width=\linewidth]{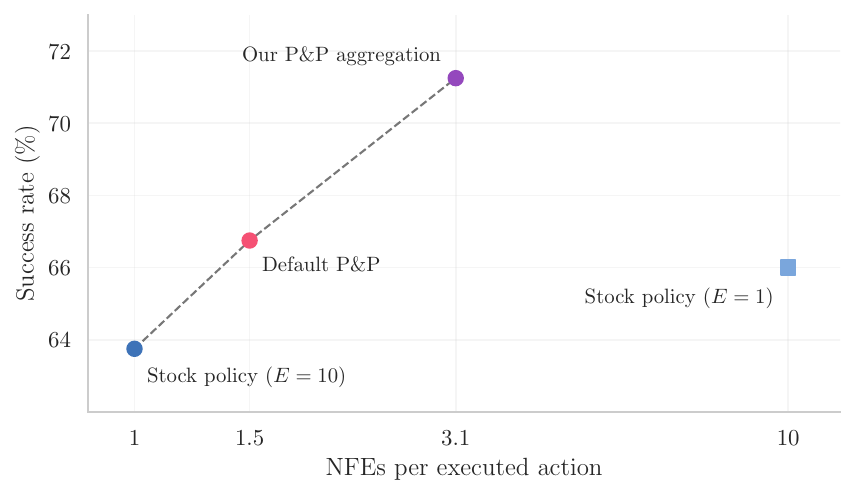}

    \end{subfigure}
    \caption{P\&P refinement is motivated as utilizing test-time computation to pull generated actions closer to a model's learned latent manifold. Top left: a schematic of the P\&P refinement process, where a sample $z_s^{(0)}$ is at an intermediate integration step . Top right: an example of stock sampling from the policy demonstrating unstable behavior for a task, resulting in failure, while P\&P-based refinement improves the stability of generated actions and ultimately enables a successful rollout. Bottom: our P\&P-based aggregation strategy more effectively leverages additional test-time computation than naively reducing the execution horizon with the stock policy or utilizing the original refinement technique \cite{jangSelfRefiningVideoSampling2026}.}
    \label{fig:pnp-comparison}
\end{figure}

\section{Introduction}
	



Learning-based robot manipulation has progressed rapidly through Vision-Language-Action models (VLAs), which combine pretrained vision-language backbones with action generators~\cite{brohanRT2VisionLanguageActionModels2023, blackP0VisionLanguageActionFlowa, nvidiaGR00TN1Open2025}. These models report high scores on popular benchmarks such as LIBERO \cite{liu2023libero}.
Most VLAs generate actions through a flow-matching process, which iteratively transforms a random noise sample into an action chunk by integrating along a learned velocity field \cite{chiDiffusionPolicyVisuomotor2024,blackP0VisionLanguageActionFlowa, nvidiaGR00TN1Open2025}. 
However, recent evidence suggests that sampled VLA trajectories are highly stochastic:
at identical simulator states, distinct noise samples can lead to markedly different trajectories and eventual task success rates \cite{yang2025steering, zhou2026test}. 

The fact that generated actions are highly variable suggests that VLAs can be improved at test time via selection, aggregation, or guidance towards an ``optimal'' action chunk that maximizes later success. Common test-time scaling methods use additional candidate generations or externally-learned critics: TACO \cite{yang2025steering} samples 32 actions and selects one using a pseudo-count estimator, while Q-Planning samples 64 actions and aggregates them using weights from a learned Q-function \cite{giridhar2026beyond}. These methods effectively improve VLA success rates, but add computational overhead. 

One potential cause of this is off-mode behavior: flow-based policies learn a distribution over actions based on finitely many expert demonstrations, which means that the learned distribution of the policy may contain incorrectly interpolated regions that do not correspond to desirable behaviors. The flow-based generation process may produce samples from these off-mode parts of the learned distribution. As such, this work investigates whether a minimal test-time correction can improve VLA success rate without the use of an external critic or the generation of many candidates, by employing test-time techniques to steer the policy towards modes in its learned distribution. 
Specifically, we adapt the \textbf{Predict-and-Perturb (P\&P)} technique for flow-based video generation models \citep{jangSelfRefiningVideoSampling2026} to VLA policies. P\&P works by iteratively ``predicting'' a clean sample (using the velocity field at an earlier timestep) and ``perturbing'' back to the designated timestep. \citet{jangSelfRefiningVideoSampling2026}  characterize this as a procedure that steers generations towards the learned data manifold, particularly towards modes, i.e., areas of high probability density. In the video paradigm, this mode-seeking behavior improves temporal stability and alignment with physical realism. 

Our main objects of study are SmolVLA-generated episodes on LIBERO, which exhibit room for improvement despite generally high success levels. While adapting P\&P, we make adjustments in cases where VLA policies do not permit tools originally used by \citet{jangSelfRefiningVideoSampling2026}. Most significantly, instead of using selective P\&P refinement on locally uncertain pixel regions, we repurpose P\&P refinement as a means for on-manifold aggregation of two action chunks.
We find that when applied to SmolVLA, P\&P refinement improves SmolVLA's success rate on LIBERO by $5.25\%$. Further, P\&P self-consistency uncertainty (a cheap signal derived from the model's velocity field under perturbation) highly correlates with episode failure (AUC $0.851$). 


\section{Background and Related Work}
\label{sec:related}

\subsection{Flow-based VLA policies and SmolVLA}

VLA policies have become a standard recipe for robot manipulation, mapping camera and language input to robot actions by scaling imitation learning over large datasets. SmolVLA \cite{shukorSmolVLAVisionLanguageActionModel2025} is a compact 450M-parameter VLA that combines a SmolVLM-2 vision-language backbone with a flow-matching transformer action expert. 
The model predicts an action chunk of $H=50$ steps at a time, along $C=7$ dimensions per step. Chunks are executed open-loop: the first $E<H$ steps are enacted before the environment is re-observed and another action chunk is generated. As such, a correction (such as P\&P refinement) applied during sampling is committed for the whole chunk before any feedback is observed. By default, the recommended configuration for SmolVLA on LIBERO uses $E=1$. 



\subsection{Inference-time correction and guidance-based steering}

Prior work has explored approaches to control the entire process of sampling from the rich, often multimodal, distribution expressed by VLAs. 

The inherent stochasticity in the sampling process can be exploited to utilize external models to perform discrete search, such as to rerank actions generated by the model \cite{nakamoto2024steering}. Another prominent approach is to augment the vector field used during the iterative integration process with auxiliary vector fields from sources such as the gradient of a learned quality function \cite{wangQVGMQGuidedValueGradient2026a, wang2026ppguide}, a learned residual actor \cite{zhu2025unified}, or an external vision-language model \cite{liu2026vls}. 

Our approach utilizes the VLA's inherent learned structure of the action distribution to perform exploration at inference time. This obviates the need for training, calibrating, or verifying an external model, and scales with the intrinsic quality of the VLA. 

\subsection{Predict-and-Perturb for Video Generation}
\label{subsec:p&p}
P\&P is a method for iteratively noising and denoising model-generated samples at predetermined flow-matching timesteps, originally applied to flow-matching models for video generation. 
Specifically, \citet{jangSelfRefiningVideoSampling2026} define the predict and perturb operators for flow timestep $s \in [0,1]$ as:
\begin{align}
  \mathcal{P}_\theta(z_s,s) &= z_s - s\,v_\theta(z_s,s) = \hat{z}_0, \label{eq:predict}\\
  \mathcal{Q}_s(\hat{z}_0,\epsilon) &= s\epsilon + (1-s)\hat{z}_0, \quad \epsilon \sim \mathcal{N}(0,I), \label{eq:perturb}
\end{align}
where $s=0,1$ denotes the posterior distribution and prior distribution, respectively (we make this choice to align with VLA flow-matching convention in Section \ref{subsec:p&p_smolvla}). 
The predict step~\eqref{eq:predict} takes one Euler extrapolation along the learned velocity field $v_\theta$ to a clean-action estimate $\hat{z}_0$; the perturb step~\eqref{eq:perturb} re-noises that estimate to the same level $s$ along the conditional path through $\hat{z}_0$.
One ``refinement step" is the composition of a prediction and a perturbation: $\mathrm{P\&P}(z_s,s) = \mathcal{Q}_s(\mathcal{P}_\theta(z_s,s), \epsilon)$. At each selected flow time $s$, we compose $K$ P\&P iterations while
holding $s$ fixed. After the final perturbation, the ordinary flow
solver resumes from the resulting latent $z_s^{(K)}$, takes its next
Euler step toward the clean endpoint $s=0$, and continues until either
the next selected refinement time or the end of sampling.

\citet{jangSelfRefiningVideoSampling2026} characterize iterative refinement as a manifold-seeking (specifically, mode-seeking) procedure, in which the latent converges toward high-density regions of the model's learned data distribution. They observe this effect within toy examples on 2D Gaussians, and similarly find that refinement reduces output variability while increasing temporal stability, resulting in more physically realistic video generation. 


The definition of self-consistency uncertainty follows. Given data with shape $C \times F \times L \times W$ (latent channels, frames, length, and width, respectively) and refinement iteration $k$, the uncertainty at each frame and pixel is:
\begin{equation}
  U_{f,h,w}^{(k)}
  =
  \frac{1}{C}\sum_{d=1}^{C}
  \left|
    \hat z_{0,d,f,h,w}^{(k+1)}
    -
    \hat z_{0,d,f,h,w}^{(k)}
  \right|.
\end{equation}
A converged trajectory yields a stable $\hat{z}_0$ under re-perturbation and hence small $U^{(k)}$.

P\&P's effectiveness is partially attributed to the video-generation paradigm; we investigate whether similar effects are possible within VLA action sampling.
For instance, it is hypothesized that P\&P's temporal stabilization properties stem from the inherent cross-frame consistency of video data. Further, 
\citet{jangSelfRefiningVideoSampling2026} improve P\&P reliability by selectively applying refinement to high-uncertainty pixel regions, thus mitigating P\&P over-correction. 
Local spatial regions do not exist within VLA action chunks; we propose a special application of P\&P that produces a similar regularization effect, in which refinement is used to aggregate stock and refined action chunks (Section \ref{subsec:p&p_agg}). 


\section{Method}
\label{sec:method}
\subsection{P\&P on SmolVLA}
\label{subsec:p&p_smolvla}
We adopt the notation of Section \ref{subsec:p&p}. Let $z_{m, s} \in \mathbb{R}^{C \times H}$ be the sampled action chunk at the $m$-th decision boundary of the episode, at flow-matching timestep $s$. Let $H$ denote action horizon (number off steps generated per chunk). The definitions of prediction and perturbation exactly follow Equations \ref{eq:predict} and \ref{eq:perturb}.

At a probed flow-matching time $s$ and refinement iteration $k$, let
$\hat z_{0,m,s,h,d}^{(k)}$ denote dimension $d$ of action step $h$
in the corresponding clean-action estimate. Analogously to the local
video uncertainty in Section~\ref{subsec:p&p}, we average the change
between consecutive clean predictions over the action dimensions, deriving the per-action-step uncertainty:
\begin{equation}
  u_{m,s,h}^{(k)}
  =
  \frac{1}{C}
  \sum_{d=1}^{C}
  \left|
    \hat z_{0,m,s,h,d}^{(k+1)}
    -
    \hat z_{0,m,s,h,d}^{(k)}
  \right|,
  \qquad h\in\{1,\ldots,H\}.
  \label{eq:vla_position_uncertainty}
\end{equation}
Thus, each P\&P comparison produces an uncertainty vector $u_{m, s, h}^{(k)} \in \mathbb{R}^{H}$ over the $H$ steps of the generated action chunk.

SmolVLA naturally generates $H=50$ actions, although only the first $E<H$ steps are executed in practice due to long-horizon chunk degradation. In our experiments, we execute only the first $E=10$ steps of each refined chunk. We therefore compute our primary uncertainty
statistics over the first $E$ action positions, excluding the unused
suffix.

Let $\mathcal{S}$ denote the set of probed flow-matching times, and let
$K_s$ be the number of P\&P refinement iterations performed at each $s\in \mathcal{S}$. To recover a single scalar uncertainty value, we first aggregate the
per-position uncertainty using a weighted average across the flow-matching times and P\&P iterations:
\begin{equation}
  u_{m,h}
  =
  \frac{1}{\sum_{s\in\mathcal{S}}K_s}
  \sum_{s\in\mathcal{S}}
  \sum_{k=1}^{K_s}
  u_{m,s,h}^{(k)}.
  \label{eq:vla_schedule_uncertainty}
\end{equation}
For an episode containing $M$ generated action chunks, the reported
episode-level uncertainty is thus the average over action chunks and executed action steps:
\begin{equation}
  \bar U_E
  =
  \frac{1}{ME}
  \sum_{m=1}^{M}
  \sum_{h=1}^{E}
  u_{m,h}.
  \label{eq:vla_episode_uncertainty}
\end{equation}



\subsection{P\&P for Refined Aggregation}
\label{subsec:p&p_agg}
In the video generation case, P\&P refinement is applied selectively to high-uncertainty pixel regions. VLA action chunks do not permit spatiotemporal selectivity; we propose an alternative method for P\&P refinement regularization.

At each decision boundary, we generate two action chunks: a stock (unrefined) chunk $a_0^{\mathrm{stock}}$ and a refined chunk $a_0^{\mathrm{ref}}$, both generated from the same initial flow noise. We first take the naive average:
$z_0 = \frac{1}{2}(a_0^{\mathrm{stock}}+a_0^{\mathrm{ref}})$. (Note that for the six continuous motion dimensions, this is an ordinary arithmetic mean. For the seventh gripper dimension, which takes on values in $\{0, 1\}$, we simply retain the stock value whenever the candidates disagree.)

However, the average of two candidate action chunks need not reside within the policy's learned action distribution. Here we use P\&P refinement to steer the average toward high-density on-manifold regions. We first slightly perturb $z_0$ to the near-clean flow time $s=0.3$:
\begin{equation}
  z_{s}^{(0)}
  =
  \mathcal{Q}_{s}(z_0,\epsilon^{(0)}),
  \qquad
  \epsilon^{(0)}\sim\mathcal{N}(0,I).
  \label{eq:aggregation_perturb}
\end{equation}

We then apply $K=3$ P\&P iterations at the fixed flow time $s=0.3$:
\begin{equation}
  z_{s}^{(k+1)}
  =
  \mathrm{P\&P}
  \left(z_{s}^{(k)},s\right),
  \qquad k=0,\ldots,K-1,
  \label{eq:aggregation_pnp}
\end{equation}
where $\mathrm{P\&P}(z_s,s)$
is the operator defined in Section~\ref{subsec:p&p}. After the final iteration, we resume the ordinary flow solver from $s\to0$.


\section{Experiments and Results}
\label{sec:result}

\subsection{Experiment Setup}
\label{subsec:setup}
We evaluate the publicly available LIBERO-finetuned SmolVLA checkpoint on the four standard LIBERO suites (LIBERO-Spatial, LIBERO-Goal, LIBERO-Object, and LIBERO-10).
Each suite contains 10 tasks with 50 numbered starting-state configurations. 
From each suite, we allocate 100 episodes (states 0-9 of each task, 400 episodes total) towards a held-out evaluation set, and 50 episodes (states 10-14) towards a validation set. We evaluate the checkpoint's unrefined policy using two configurations: $E=1$ executed actions per chunk generation (the recommended default configuration) and $E=10$. 
Unless stated otherwise, we use $H=50$, $C=7$, and ten Euler integration steps. 

In total, six configurations are compared on this 400-episode evaluation set: the stock and refined policies are each evaluated with $E\in\{1, 10\}$.
The refined steps are $s=0.9,0.8, 0.7$ (corresponding to Euler steps $i=1, 2, 3$) with $K_{0.9}=3, K_{0.8}=K_{0.7}=1$ refinement iterations.
Two additional stock/refinement aggregation
configurations (described in Section \ref{subsec:p&p_agg}) are included: a refined aggregation arm and a naive average baseline, each using $E=10$. We report overall success rates for each configuration, along with per-suite success rate and episode-level transition comparisons for select configurations. Finally, we evaluate whether high episode-level $U_{10}$ (defined in Sec. \ref{subsec:p&p_smolvla}) during P\&P refinement correlates highly with task failure.

\subsection{Success Rate and Uncertainty Analysis}
\label{subsec:sr_uncertainty}

\begin{figure}[h]
  \centering
  \includegraphics[width=0.8\linewidth]{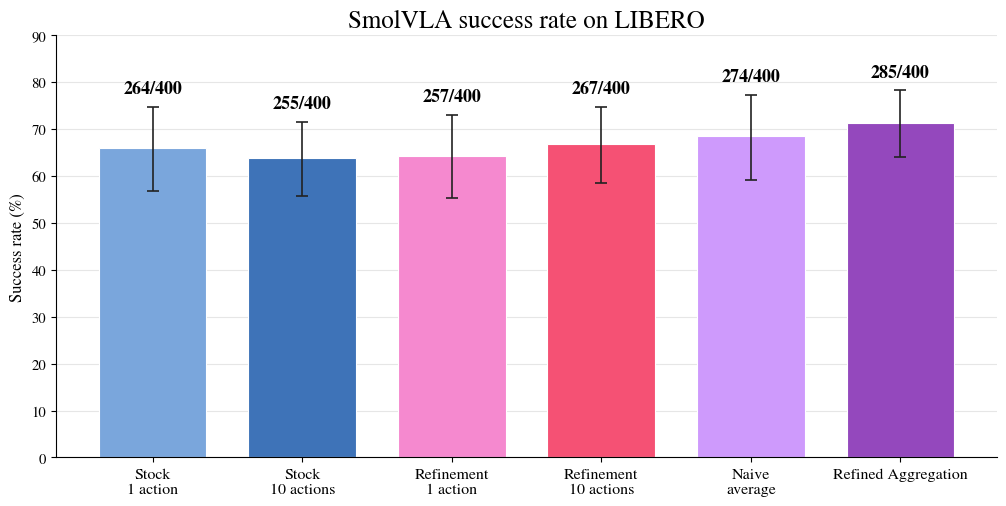}
  \caption{Success rate of SmolVLA variants over 400 matched LIBERO eval episodes. Stock sampling and P\&P refinement are each evaluated with execution horizons of 1, 10. Naive and refined averages both aggregate stock and refined action chunks and execute the first 10. Labels report number of successful episodes. Error bars show 95\% percentile task-cluster bootstrap intervals over the $40$ tasks, where all evaluated initial states for a sampled task are retained together.} 
  \label{fig:pooled_sr}
\end{figure}

\begin{figure}[h]
  \centering
  \includegraphics[width=0.85\linewidth]{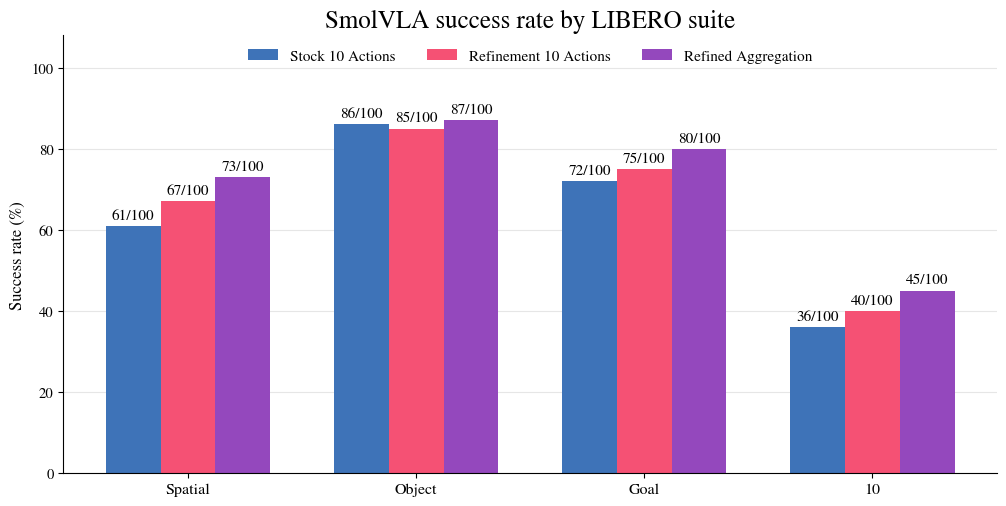}
  \caption{Per-suite success rate of stock, refined, and refined aggregation policies over 400 matched LIBERO episodes. Each configuration uses an action horizon of $E=10$.}
  \label{fig:suite_sr}
\end{figure}

\begin{figure}[h]
  \centering
  \includegraphics[width=0.9\linewidth]{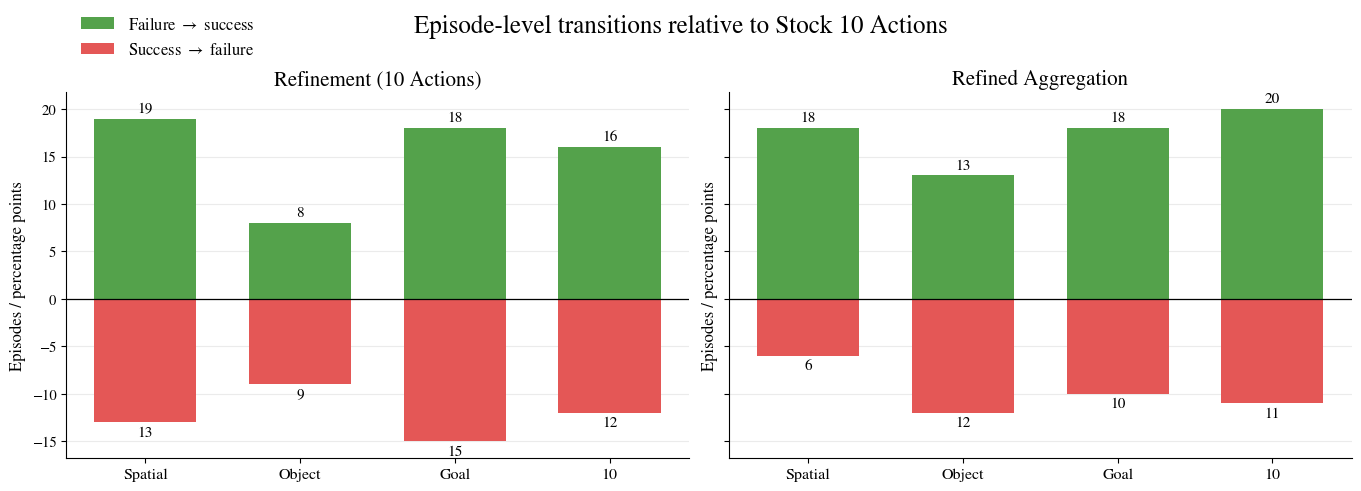}
  \caption{Per-suite episode transitions of P\&P refinement (10 actions) and refined aggregation relative to stock SmolVLA. Green bars denote recovered episodes (Failure $\to$ Success), red bars denote degraded episodes (Success $\to$ Failure).}
  \label{fig:episode_trans}
\end{figure}

\begin{figure}[h]
  \centering
  \includegraphics[width=0.6\linewidth]{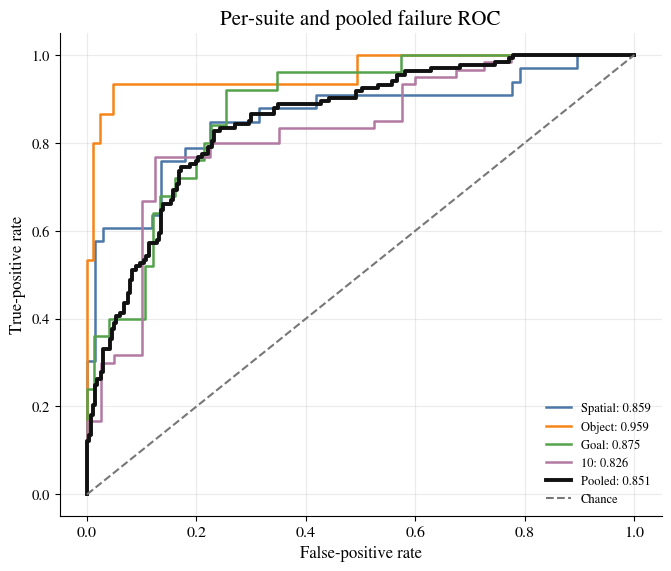}
  \caption{Failure-prediction AUROC curves for episode-level uncertainty $U_{10}$ (defined in Section \ref{subsec:p&p_smolvla}) during ten-action refinement. Colored curves represent LIBERO suite: Spatial, Object, Goal, 10. Pooled AUC across the 400-episode subset reaches 0.851.}
  \label{fig:auc}
\end{figure}

Figure \ref{fig:pooled_sr} compares all six configurations on the 400-episode LIBERO subset. Notably, refined aggregation (which uses an $E=10$ execution horizon) yields a $5.25\%$ increase over the highest-scoring stock configuration ($E=1$), and a $7.5\%$ increase over the corresponding $E=10$ stock configuration. Naive averaging also improves performance (yielding the second-highest success rate, $2.5\%$ above stock with $E=1$), but falls short of refined aggregation by $11$ episodes. This additional improvement is consistent with the manifold-seeking motivation described in 
Section \ref{subsec:p&p_agg}, although future work should distinguish this mechanism from alternative explanations (such as action smoothing) by directly inspecting the geometry of action chunks.

Also of note is the effect of P\&P as the action horizon varies. Refinement of action chunks is seemingly helpful when later action steps are executed ($E=10$ improves from $255/400$ (stock) $\to 267/400$ (refined)), perhaps mitigating the common failure mode of long-horizon degradation. Conversely, refinement is not helpful in the case where $E=1$, as performance degrades ($264\to257$), likely from cases where P\&P overcorrects otherwise successful trajectories.

Figure \ref{fig:suite_sr} shows the per-suite success rates of stock (10 actions), refinement (10 actions), and refined aggregation. Meanwhile, Figure \ref{fig:episode_trans} reports an episode-level transition analysis of refinement (10 actions) and refined aggregation relative to the stock. These figures demonstrate refined aggregation serving its intended purpose of P\&P regularization: the ratio of recovered to degraded episodes during refined aggregation is improved across the board, most substantially on LIBERO-Spatial ($19/13 \to 18/6$) and LIBERO-10 ($16/12\to 20/10$). Within these same suites, success rate gain substantially increases (from $6\% \to 12\%$ (LIBERO-Spatial) and $4\% \to 9\%$ (LIBERO-10)).






Finally, Figure \ref{fig:auc} evaluates whether episode-level uncertainty $U_{10}$ (Section \ref{subsec:p&p_smolvla}) distinguishes successful from failed trajectories during refinement with $E=10$. Assigning failed episodes to be the positive class, AUC reaches 0.851 on the pooled 400-episode subset and ranges from $0.826$ (LIBERO-10) to $0.959$ (LIBERO-Object). This establishes a strong correlation between episode-level uncertainty and task failure. However, since episode-level uncertainty must  be computed retrospectively over completed episodes, further work may explore utilizing uncertainty during an incomplete episode for online failure prediction. 



\section{Conclusion}
\label{sec:conclusion}
We adapt Predict-and-Perturb, originally developed for self-refining video generation, to flow-matching VLA policies. Along with P\&P refinement, we propose refined aggregation, in which the average of a stock and refined chunk is further refined at a late flow time $s=0.3$ in order to steer towards higher-density regions of the action distribution. Across a 400-episode LIBERO evaluation set, our method improves upon the stock policy by $5.25\%$. We additionally find that episode-level uncertainty is
strongly associated with failure, achieving a pooled failure AUC of $0.851$. Together, these results suggest that a flow policy's own refinement dynamics
can provide useful test-time intervention.

\paragraph{Future Work.} Our investigation is limited to the SmolVLA LIBERO checkpoint and 400-episode LIBERO subset. Further experiments could extend this investigation to larger models (such as $\pi_{0.5}$, which approaches saturation on LIBERO), evaluate using repeated-seed results, and assess whether refinement assists in real-world observations and difficult, out-of-distribution benchmarks such as LIBERO-PRO. 

Further work should closely analyze whether chunk-level uncertainty can be purposed as a useful online signal for selective refinement. 

Another promising direction lies in utilizing P\&P for the purpose of training a lightweight Q-corrector. Combining learned guidance towards correct behavior with P\&P's model-internal
self-consistency signal towards mode-seeking behavior, and exploration of refined trajectories, may enable more selective and reliable deployment-time correction.


\clearpage
\acknowledgments{We thank the CoRL anonymous reviewer for constructive feedback on improving statistical analysis, compute-matched analysis, and discussion of method limitations.}


\bibliography{refs}  

@article{blackP0VisionLanguageActionFlowa,
  title = {{$\pi$}0: {{A Vision-Language-Action Flow Model}} for {{General Robot Control}}},
  author = {Black, Kevin and Brown, Noah and Driess, Danny and Esmail, Adnan and Equi, Michael and Finn, Chelsea and Fusai, Niccolo and Groom, Lachy and Hausman, Karol and Ichter, Brian and Jakubczak, Szymon and Jones, Tim and Ke, Liyiming and Levine, Sergey and {Li-Bell}, Adrian and Mothukuri, Mohith and Nair, Suraj and Pertsch, Karl and Shi, Lucy Xiaoyang and Tanner, James and Vuong, Quan and Walling, Anna and Wang, Haohuan and Zhilinsky, Ury},
  langid = {english}
}

@misc{brohanRT2VisionLanguageActionModels2023,
  title = {{{RT-2}}: {{Vision-Language-Action Models Transfer Web Knowledge}} to {{Robotic Control}}},
  shorttitle = {{{RT-2}}},
  author = {Brohan, Anthony and Brown, Noah and Carbajal, Justice and Chebotar, Yevgen and Chen, Xi and Choromanski, Krzysztof and Ding, Tianli and Driess, Danny and Dubey, Avinava and Finn, Chelsea and Florence, Pete and Fu, Chuyuan and Arenas, Montse Gonzalez and Gopalakrishnan, Keerthana and Han, Kehang and Hausman, Karol and Herzog, Alexander and Hsu, Jasmine and Ichter, Brian and Irpan, Alex and Joshi, Nikhil and Julian, Ryan and Kalashnikov, Dmitry and Kuang, Yuheng and Leal, Isabel and Lee, Lisa and Lee, Tsang-Wei Edward and Levine, Sergey and Lu, Yao and Michalewski, Henryk and Mordatch, Igor and Pertsch, Karl and Rao, Kanishka and Reymann, Krista and Ryoo, Michael and Salazar, Grecia and Sanketi, Pannag and Sermanet, Pierre and Singh, Jaspiar and Singh, Anikait and Soricut, Radu and Tran, Huong and Vanhoucke, Vincent and Vuong, Quan and Wahid, Ayzaan and Welker, Stefan and Wohlhart, Paul and Wu, Jialin and Xia, Fei and Xiao, Ted and Xu, Peng and Xu, Sichun and Yu, Tianhe and Zitkovich, Brianna},
  year = 2023,
  month = jul,
  number = {arXiv:2307.15818},
  eprint = {2307.15818},
  primaryclass = {cs},
  publisher = {arXiv},
  doi = {10.48550/arXiv.2307.15818},
  urldate = {2026-05-06},
  archiveprefix = {arXiv}
}

@misc{chiDiffusionPolicyVisuomotor2024,
  title = {Diffusion {{Policy}}: {{Visuomotor Policy Learning}} via {{Action Diffusion}}},
  shorttitle = {Diffusion {{Policy}}},
  author = {Chi, Cheng and Xu, Zhenjia and Feng, Siyuan and Cousineau, Eric and Du, Yilun and Burchfiel, Benjamin and Tedrake, Russ and Song, Shuran},
  year = 2024,
  month = mar,
  number = {arXiv:2303.04137},
  eprint = {2303.04137},
  primaryclass = {cs},
  publisher = {arXiv},
  doi = {10.48550/arXiv.2303.04137},
  urldate = {2026-05-06},
  archiveprefix = {arXiv}
}

@misc{jangSelfRefiningVideoSampling2026,
  title = {Self-{{Refining Video Sampling}}},
  author = {Jang, Sangwon and Ki, Taekyung and Jo, Jaehyeong and Xie, Saining and Yoon, Jaehong and Hwang, Sung Ju},
  year = 2026,
  month = jan,
  number = {arXiv:2601.18577},
  eprint = {2601.18577},
  primaryclass = {cs},
  publisher = {arXiv},
  doi = {10.48550/arXiv.2601.18577},
  urldate = {2026-05-03},
  archiveprefix = {arXiv}
}

@misc{nvidiaGR00TN1Open2025,
  title = {{{GR00T N1}}: {{An Open Foundation Model}} for {{Generalist Humanoid Robots}}},
  shorttitle = {{{GR00T N1}}},
  author = {NVIDIA and Bjorck, Johan and Casta{\~n}eda, Fernando and Cherniadev, Nikita and Da, Xingye and Ding, Runyu and Fan, Linxi "Jim" and Fang, Yu and Fox, Dieter and Hu, Fengyuan and Huang, Spencer and Jang, Joel and Jiang, Zhenyu and Kautz, Jan and Kundalia, Kaushil and Lao, Lawrence and Li, Zhiqi and Lin, Zongyu and Lin, Kevin and Liu, Guilin and Llontop, Edith and Magne, Loic and Mandlekar, Ajay and Narayan, Avnish and Nasiriany, Soroush and Reed, Scott and Tan, You Liang and Wang, Guanzhi and Wang, Zu and Wang, Jing and Wang, Qi and Xiang, Jiannan and Xie, Yuqi and Xu, Yinzhen and Xu, Zhenjia and Ye, Seonghyeon and Yu, Zhiding and Zhang, Ao and Zhang, Hao and Zhao, Yizhou and Zheng, Ruijie and Zhu, Yuke},
  year = 2025,
  month = mar,
  number = {arXiv:2503.14734},
  eprint = {2503.14734},
  primaryclass = {cs},
  publisher = {arXiv},
  doi = {10.48550/arXiv.2503.14734},
  urldate = {2026-05-06},
  archiveprefix = {arXiv}
}

@misc{shukorSmolVLAVisionLanguageActionModel2025,
  title = {{{SmolVLA}}: {{A Vision-Language-Action Model}} for {{Affordable}} and {{Efficient Robotics}}},
  shorttitle = {{{SmolVLA}}},
  author = {Shukor, Mustafa and Aubakirova, Dana and Capuano, Francesco and Kooijmans, Pepijn and Palma, Steven and Zouitine, Adil and Aractingi, Michel and Pascal, Caroline and Russi, Martino and Marafioti, Andres and Alibert, Simon and Cord, Matthieu and Wolf, Thomas and Cadene, Remi},
  year = 2025,
  month = jun,
  number = {arXiv:2506.01844},
  eprint = {2506.01844},
  primaryclass = {cs},
  publisher = {arXiv},
  doi = {10.48550/arXiv.2506.01844},
  urldate = {2026-05-06},
  archiveprefix = {arXiv}
}

@misc{wangQVGMQGuidedValueGradient2026a,
  title = {Q-{{VGM}}: {{Q-Guided Value-Gradient Matching}} for {{Flow-Matching VLA Policies}}},
  shorttitle = {Q-{{VGM}}},
  author = {Wang, Ziqian and Sun, Jiayu and Mao, Xingjian and Wang, Minqian and Mu, Yao},
  year = 2026,
  month = jun,
  number = {arXiv:2606.08015},
  eprint = {2606.08015},
  primaryclass = {cs.RO},
  publisher = {arXiv},
  doi = {10.48550/arXiv.2606.08015},
  urldate = {2026-08-21},
  archiveprefix = {arXiv}
}

@article{yang2025steering,
  title={Steering vision-language-action models as anti-exploration: A test-time scaling approach},
  author={Yang, Siyuan and Zhang, Yang and He, Haoran and Pan, Ling and Li, Xiu and Bai, Chenjia and Li, Xuelong},
  journal={arXiv preprint arXiv:2512.02834},
  year={2025}
}

@article{zhou2026test,
  title={Test-Time Gradient Guidance of Flow Policies in Reinforcement Learning},
  author={Zhou, Zhiyuan and Peng, Andy and Xu, Charles and Li, Qiyang and Springenberg, Tobias and Frans, Kevin and Levine, Sergey},
  journal={arXiv preprint arXiv:2606.11087},
  year={2026}
}

@article{giridhar2026beyond,
  title={Beyond imitation: Self-improving robot policies via off-policy Q-planning},
  author={Giridhar, Varun and Khandelwal, Anant and Collins, Jeremy A and Georgiev, Ignat and Garg, Animesh},
  journal={arXiv preprint arXiv:2608.21204},
  year={2026}
}

@article{liu2023libero,
  title={Libero: Benchmarking knowledge transfer for lifelong robot learning},
  author={Liu, Bo and Zhu, Yifeng and Gao, Chongkai and Feng, Yihao and Liu, Qiang and Zhu, Yuke and Stone, Peter},
  journal={Advances in Neural Information Processing Systems},
  volume={36},
  pages={44776--44791},
  year={2023}
}

@article{zhu2025unified,
  title={Unified Latent Steering and Residual Refinement for Online Improvement of Diffusion Policy Models},
  author={Zhu, Zhengbang and Li, Ziyan and Yuan, Xiu and Zhang, Hanbo and Liu, Yuxiao and Zhang, Chongjie and Yu, Yong and Zhang, Weinan and Liu, Minghuan},
  year={2025}
}

@article{liu2026vls,
  title={Vls: Steering pretrained robot policies via vision-language models},
  author={Liu, Shuo and Singh, Ishneet Sukhvinder and Xu, Yiqing and Duan, Jiafei and Krishna, Ranjay},
  journal={arXiv preprint arXiv:2602.03973},
  year={2026}
}

@article{wang2026ppguide,
  title={Ppguide: Steering diffusion policies with performance predictive guidance},
  author={Wang, Zixing and Jha, Devesh K and Qureshi, Ahmed H and Romeres, Diego},
  journal={arXiv preprint arXiv:2603.10980},
  year={2026}
}

@article{nakamoto2024steering,
  title={Steering your generalists: Improving robotic foundation models via value guidance},
  author={Nakamoto, Mitsuhiko and Mees, Oier and Kumar, Aviral and Levine, Sergey},
  journal={arXiv preprint arXiv:2410.13816},
  year={2024}
}

\clearpage
\appendix

\section{Reproducibility}
All experiments use the publicly available
\texttt{HuggingFaceVLA/smolvla\_libero} checkpoint without additional
fine-tuning. The rollouts included in the main text are evaluated using batch size 8 and a single NVIDIA L4 GPU. All tasks included in the 400-episode evaluation set and the 200-episode validation set stem from the 2000-episode LIBERO test split, which includes 50 numbered initial states for each task.

\section{NFE Cost and Matching}
\subsection{Neural function evaluation calculations}
\label{nfe_calcs}

In \autoref{fig:pnp-comparison}, we compare neural function evaluations (NFEs) per executed action for our approaches to sampling from the stock policy. Here, we present the calculations used to generate that plot. 

The stock policy uses ten Euler steps. Default P\&P adds $K_{0.9}+K_{0.8}+K_{0.7}=3+1+1$ predictions. Our P\&P aggregation strategy generates two chunks, performs three fixed-time iterations at and completes three Euler steps. Table~\ref{tab:nfe} summarizes these costs. NFEs per executed action are obtained by dividing by the execution horizon $E$.

\begin{table}[htbp]
\centering
\caption{Action-generation cost in NFEs.}
\label{tab:nfe}
\begin{tabular}{lccc}
\hline
Method & $E$ & NFEs/chunk & NFEs/action \\
\hline
Stock policy         & 1  & 10 & 10 \\
Stock policy         & 10 & 10 & 1 \\
Default P\&P         & 10 & $10+3+1+1=15$ & 1.5 \\
Our P\&P aggregation & 10 & $10+15+3+3=31$ & 3.1 \\
\hline
\end{tabular}
\end{table}

\subsection{Compute-Matched Baselines}

\begin{figure}[h]
  \centering
  \includegraphics[width=0.85\linewidth]{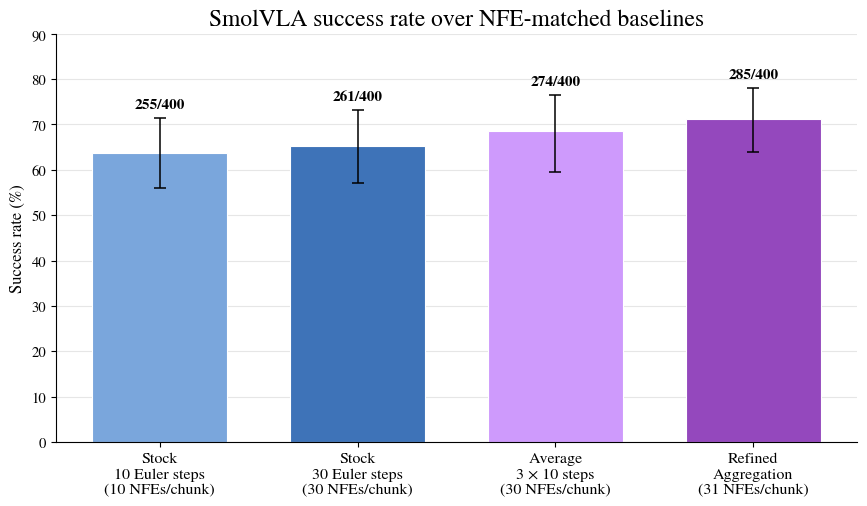}
  \caption{Success rate of NFE-matched baselines, along with stock arm using 10 Euler steps shown for success rate reference. From left to right: stock (10 Euler steps), stock (30 Euler steps), 3-candidate average, refined aggregation. Error bars show 95\% percentile task-cluster bootstrap intervals over the $40$ tasks, where all evaluated initial states for a sampled task are retained together.}
  \label{fig:compute_match}
\end{figure}

To control for the additional inference required by P\&P aggregation (31 NFEs/chunk by Section \ref{nfe_calcs}), we compare P\&P against two critic-free baselines with approximately equal NFE budgets. The first baseline allocates $30$ NFEs per chunk by increasing the flow integration horizon from $10$ to $30$ Euler steps. The second generates three independent stock action candidates, each using $10$ Euler steps, and aggregates them via averaging. All methods use $E=10$ actions per query and are evaluated on the same $400$ LIBERO episodes mentioned in Section \ref{subsec:setup}.

Figure \ref{fig:compute_match} shows that $30$ Euler steps only marginally improves upon the $10$-step baseline ($255\to261$ successes). Three-way averaging performs equally to naive P\&P aggregation (Figure \ref{fig:pooled_sr}) and $2.75\%$ lower than refined aggregation. Future work could extend compute analyses to per-chunk latency and total simulation time.





\end{document}